\pdfoutput=1

\documentclass[11pt]{article}

\usepackage{ACL2023}

\usepackage{times}
\usepackage{latexsym}

\usepackage[T1]{fontenc}

\usepackage[utf8]{inputenc}

\usepackage{microtype}

\usepackage{inconsolata}
\usepackage{booktabs}
\usepackage{multirow}
\usepackage{caption}
\usepackage{subcaption}
\usepackage{graphicx}

\title{Making More of Little Data: Improving Low-Resource Automatic Speech Recognition Using Data Augmentation}

\author{
\textbf{Martijn Bartelds}\textsuperscript{$1$}
\quad \textbf{Nay San}\textsuperscript{$2$} \quad \textbf{Bradley McDonnell}\textsuperscript{$3$} \\ \textbf{Dan Jurafsky}\textsuperscript{$2$} \quad \textbf{Martijn Wieling}\textsuperscript{$1$} \\
\textsuperscript{$1$}University of Groningen\quad
\textsuperscript{$2$}Stanford University\quad
\textsuperscript{$3$}University of Hawai'i at M\=anoa\\
\texttt{m.bartelds@rug.nl}
}

\begin{document}
\maketitle
\begin{abstract}
The performance of automatic speech recognition (ASR) systems has advanced substantially in recent years, particularly for languages for which a large amount of transcribed speech is available.
Unfortunately, for low-resource languages, such as minority languages, regional languages or dialects, ASR performance generally remains much lower.
In this study, we investigate whether data augmentation techniques could help improve low-resource ASR performance, focusing on four typologically diverse minority languages or language variants (West Germanic: Gronings, West-Frisian; Malayo-Polynesian: Besemah, Nasal).
For all four languages, we examine the use of self-training, where an ASR system trained with the available human-transcribed data is used to generate transcriptions, which are then combined with the original data to train a new ASR system.
For Gronings, for which there was a pre-existing text-to-speech (TTS) system available, we also examined the use of TTS to generate ASR training data from text-only sources.
We find that using a self-training approach consistently yields improved performance (a relative WER reduction up to 20.5\% compared to using an ASR system trained on 24 minutes of manually transcribed speech).
The performance gain from TTS augmentation for Gronings was even stronger (up to 25.5\% relative reduction in WER compared to a system based on 24 minutes of manually transcribed speech).
In sum, our results show the benefit of using self-training or (if possible) TTS-generated data as an efficient solution to overcome the limitations of data availability for resource-scarce languages in order to improve ASR performance.
\end{abstract}

\section{Introduction}
Self-supervised learning (SSL) enables speech representation learning without the need for (manually) labeled data.
Although this approach is very effective, pre-training an SSL model is costly.
This cost (e.g., training time, resources, and memory) increases with the number of languages added to the model.
Furthermore, transferring information across languages, or extending a pre-trained model to new data or to a different domain is computationally expensive, and catastrophic forgetting may occur \cite{c-forgetting}.
To alleviate this, SSL models are therefore often fine-tuned on the target task with target domain data.
For the task of automatic speech recognition (ASR), fine-tuning approaches generally require less data, but training ASR systems that perform well for languages with very little data remains challenging.
This leads to (digitally) underrepresented communities and domains such as minority languages, regional languages and dialects not profiting sufficiently from most recent technological advancements.

Recent studies explored fine-tuning of pre-trained self-supervised models for ASR using speech from low-resource languages (e.g.,~\citealt{coto-solano-etal-2022-development, guillaume-etal-2022-fine}), and difficulties of modeling resource-scarce languages and dialects were acknowledged in previous work \citep{vanesch-2022}.
It remains an open question to what extent model performance is dependent on the amount of fine-tuning data and the type of language, when the total amount of available data for a language is limited.
Having a better understanding of how limited training data affects model performance paves the way for creating meaningful speech technology for a wider range of languages.

In this paper, we fine-tune pre-trained SSL models for ASR using varying amounts of data from four typologically diverse minority languages or language variants: Gronings, West-Frisian, Besemah and Nasal, which have a limited amount of data available.
We specifically investigate whether data augmentation approaches can be used to generate additional training data to improve the performance of these models, particularly when very little resources are available.
By using data from (ongoing) language documentation projects, we evaluate a real-world use of our experimental setup.

Previous work describes the benefits of data augmentation by adopting a self-training approach, which generates labels (i.e.~transcriptions) for unlabeled speech (e.g., \citealt{iterative-pseudo, self-complementary, self-end-to-end, xlst, cont-pseudo, magic-dust, pseudo-multiling}).
Various self-training methods are proposed, including iterative approaches, decoding with an external (text-based) language model, or filtering approaches that improve the quality of the generated labels.
However, limited conclusions can be drawn from these works on the effectiveness of self-training in a very low-resource, real-world setting, as these studies either use datasets with more than 10 hours of data (which may not be available for very small languages), only considered modeling English, or reported average performance over a set of languages that strongly varied in terms of training data size.
We therefore complement this work by investigating the benefits of self-training for four typologically different, true low-resource languages.
To this end, we use a standard self-training approach to evaluate the potential benefit of a simple system in a real-world setup, which nevertheless yields substantial performance improvements (relative word-error-rate (WER) reductions up to 20.5\%).

In addition to self-training, several studies (e.g., \citealt{rosenberg-tts, 9053139, rossenbach-tts}) reported on augmenting the training data with synthetic speech generated using a text-to-speech (TTS) system.
For this reason, we also examine whether this approach is useful in our low-resource setup.
We recognize that not all very low-resource languages may have sufficient amounts of data available for TTS development, and we therefore only generate synthetic training examples for Gronings, one of the four low-resource languages in our dataset that has an existing TTS system available.
We show the benefit (i.e.~up to 25.5\% relative reduction in WER) of augmenting the training data by using an existing TTS system, and analyze the effect of adding different amounts of synthetic speech on the model performance.
Our datasets, code, and newly trained models are publicly available.\footnote{\url{https://github.com/Bartelds/asr-augmentation}}

\section{Data}
As indicated, we use transcribed speech from Gronings, West-Frisian, Besemah, and Nasal.
For the latter two minority languages, only four hours of manually transcribed speech data are available.
For all language varieties, we therefore limit the amount of manually transcribed speech data to four hours.
We divide each dataset into 80\% for training, 10\% for development and 10\% for testing.
The development and test sets therefore include approximately 24 minutes of speech, and the training set contains approximately 3.2 hours of transcribed speech.
In line with \citet{wei:2022}, we allow for speaker overlap between the sets due to the limited number of speakers per language variant, as they found that it had limited effects on the performance of ASR models.
All data have been anonymized by assigning recordings a random identifier, and no other meta-information that could be used for identifying the speakers were collected or extracted.
We obtained consent from the communities to publicly release the datasets for Gronings, Besemah, and Nasal.
The West-Frisian data can be obtained by emailing the authors (ISLRN: 340-994-352-616-4).

\subsection{Gronings and West-Frisian}
Gronings is a Low-Saxon language variant that is spoken in the province of Groningen, which is located in the northern part of the Netherlands.
Within this language variant, there is regional lexical, grammatical and acoustic variation.
We use data from an ongoing language documentation project that aims to record the speech of all variants of Gronings.
To date, read-aloud speech from three speakers has been recorded (two female speakers and one male speaker) for three different variants, namely Hogelandsters, Oldambtsters, and Westerkwartiers.
This data, consisting of almost 14 hours of transcribed speech data, is included in this study.
From these 14 hours, four hours of manually transcribed speech was extracted for training, development and testing.
The remaining data was partly used for generating additional training data.  
The 2,130 transcribed recordings in this dataset, comprised of book texts and corresponding recordings, have an average duration of 6.8 seconds (SD: 4.9).
We normalized the transcriptions by excluding all characters that do not occur in the Gronings alphabet.\footnotemark~In addition, we also include transcribed speech data from three different speakers (two female speakers and one male speaker), yielding a total of 19 minutes of speech data.
This data was extracted from the publicly available dataset provided by \citet{san-bartelds}.
These recordings have a mean duration of 3.5 seconds (SD: 1.3).
We only use this subset of data for out-of-domain testing. 

West-Frisian is the second official language of the Netherlands and is spoken in the province of Friesland, which is also located in the northern part of the Netherlands.
For this study, we extracted four (out of eight) hours of transcribed speech data from the FAME! ASR corpus \citep{yilmaz2017longitudinal} that contains radio and television speech from Dutch-Frisian bilinguals.
The extracted dataset includes 4,919 transcribed speech samples from 277 speakers (68 female, 199 male speakers, and 10 unknown) with an average duration of 2.9 seconds (SD: 0.7).
We removed all characters from the transcripts that are not part of the West-Frisian alphabet \citep{ylmaz16b_interspeech}.

\subsection{Besemah and Nasal}
Besemah and Nasal are two Austronesian languages that are spoken in southern Sumatra, Indonesia.
For both languages, approximately 45 hours of informal conversation data were collected through fieldwork.
For each language, four hours of conversational data have been transcribed, which are used in this study.
For Besemah, there are 7,835 transcribed utterances from 46 speakers (30 female speakers and 16 male speakers) with an average sample length of 1.8 seconds (SD: 0.3).
The Nasal dataset contains 7,672 transcribed utterances from 40 speakers (15 female speakers and 25 male speakers) with an average duration of 3.9 seconds (SD: 0.3). 
We normalized all transcriptions to the working orthographies developed for Besemah and Nasal as part of ongoing collaborative language documentation projects. \footnotetext{\url{https://woordwaark.nl/spelling.pdf}}

\section{Methods}
We fine-tune the pre-trained multilingual \texttt{XLS-R} model with 317 million parameters on different amounts of training data from the four languages in our dataset \citep{xls-r}.
Note that we chose the smallest publicly available pre-trained \texttt{XLS-R} model to minimize the computational requirements needed for (reproducing) this study.
\texttt{XLS-R} is pre-trained on approximately 436,000 hours of speech in 128 different languages.
This data was collected from a variety of sources, including parliamentary speech (372,000 hours in 23 European languages), read speech from Multilingual Librispeech (44,000 hours in eight European languages) and Common Voice (7,000 hours in 60 languages), speech from YouTube from the VoxLingua107 corpus (6,600 hours in 107 languages), and conversational telephone speech from the BABEL corpus (approximately 1,000 hours in 17 African and Asian languages).
The majority of the training data is from Indo-European languages (87\%), and the language that is most represented is English (roughly 70,000 hours).
While the model does include a small portion of West-Frisian data (i.e.~15 hours), this is not the case for Gronings, Besemah, and Nasal.

The architecture and pre-training objective of \texttt{XLS-R} are similar to those of wav2vec 2.0 \citep{NEURIPS2020_92d1e1eb}.
The model is trained as a single end-to-end system, and consists of a convolutional encoder, a quantizer, and a 24-layer Transformer model.
Speech representations are learned through a contrastive task that is applied to the quantized encoder representations.
After pre-training, the model can be fine-tuned for speech recognition using transcribed speech.
A linear projection is added on top of the Transformer network to predict characters from the transcriptions using connectionist temporal classification (CTC; \citealt{graves2006connectionist}).

We include a multilingual model in our study, because previous work showed that multilingual pre-training transfers well to low-resource languages (e.g., \citealt{bartelds-wieling-2022-quantifying, magic-dust}).
We experimented with fine-tuning other models (for example the Dutch wav2vec 2.0 model included by \citealt{bartelds-wieling-2022-quantifying}), but preliminary results showed that \texttt{XLS-R} was superior.
The hyperparameters of our fine-tuning experiments follow those reported in \citet{NEURIPS2020_92d1e1eb} for comparable data sizes, except for the learning rate, which we tune on the basis of the development data by evaluating the following range: $[5\mathrm{e}{-4}, 1\mathrm{e}{-4}, 5\mathrm{e}{-5}, 1\mathrm{e}{-5}]$.
In addition, we reduce the batch size and use gradient accumulation to make sure our experiments run on limited compute hardware (i.e.~a single Nvidia 40 GB A100 GPU).
We evaluate the fine-tuned models in terms of word error rate (WER), which is a commonly used evaluation metric based on the number of substitutions, deletions, and additions between two transcripts, and report performance on the test set using the fine-tuned model checkpoint that has the lowest WER on the validation set.

Additionally, we investigate whether it is beneficial to further pre-train the \texttt{XLS-R} model using limited data and computational hardware before fine-tuning the model for ASR.
As pre-training is computationally expensive, we only evaluate the performance on Gronings, for which we perform the broadest range of experiments.
Specifically, we pre-train on the four hours of Gronings training data with the test set samples removed for 100,000 steps and use a learning rate of $1\mathrm{e}{-5}$, which was selected after briefly experimenting with a range of learning rates that we evaluated on the validation set.
Similar to the fine-tuning experiments, we use gradient accumulation and a small batch size.

The total computational budget for this study is about 390 hours on a 40 GB A100 GPU (160 fine-tuning runs of roughly 2 hours each, and pre-training runs of roughly 70 hours).
We perform all experiments using the HuggingFace Transformers library, version 4.24.0 \citep{wolf-etal-2020-transformers}.

\section{Experimental Setup}
For each of the languages, we use varying amounts of training data for fine-tuning the multilingual \texttt{XLS-R} model.~Additionally, for Gronings, we also fine-tune the \texttt{XLS-R} model that is further pre-trained on Gronings.
For all experiments, we start from the full training dataset of 192 minutes (80\% of four hours), and divide this set repeatedly into smaller subsets until reaching roughly 20 minutes (50\% of each split).
Consequently, we have training sets of 192, 96, 48 and 24 minutes, respectively.

\begin{figure*}[ht!]
    \centering
    \includegraphics[width=0.79\textwidth]{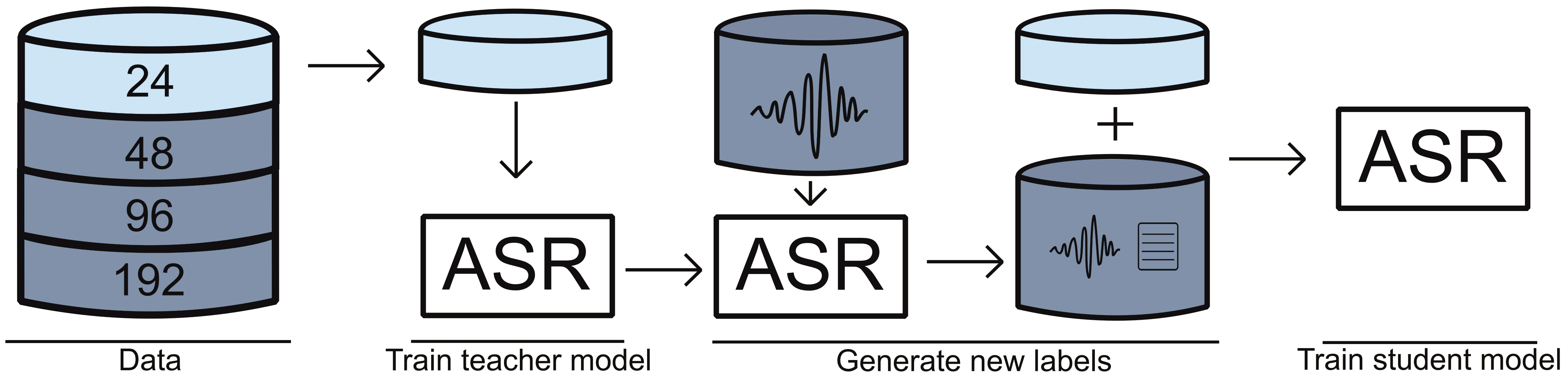}
    \caption{Visualization of the self-training approach where a teacher model is fine-tuned on manually transcribed data and subsequently used to transcribe unlabeled speech. A student model is then fine-tuned on the combined datasets.}
    \label{fig:method:st}
\end{figure*}

In the self-training approach, we fine-tune the pre-trained \texttt{XLS-R} models on one of the subsets of data (i.e.~24, 48, or 96 minutes) as the initial step.
We regard this model as the teacher model, which is then used to transcribe the remaining portion of speech data from the full training data (i.e.~without the labels).
The resulting automatically transcribed data, in conjunction with the original labeled data, is subsequently used to fine-tune a second model, referred to as the student model, which ideally outperforms the teacher model.~This approach is shown in Figure~\ref{fig:method:st}.
For example, we fine-tune a \texttt{XLS-R} teacher model on 24 minutes of manually transcribed speech data and use this model to label the remaining 168 minutes of speech data contained in the full training set.~The combined data (e.g., 24 minutes of natural speech with correct labels and 168 minutes of automatically transcribed speech obtained through self-training) are subsequently used to fine-tune a new student model.
We apply this procedure to each of the three training splits to investigate in which cases self-training may be beneficial in a low-resource setting.
Our decoding procedure does not use an external language model (LM) due to the limited availability of text-based training materials for all languages, and also to ensure a fair comparison between languages.
This is supported by previous work that found no improvement in speech recognition performance when limited amounts of textual data are available for LM training \citep{san-etal-2023-leveraging}.

Note that in addition to the self-training approach, preliminary experiments were conducted with other data augmentation techniques (following \citealt{sriram2022wav2vecaug}).
Specifically, we experimented with adding noise to the speech signal, raising or lowering the pitch of the speaker, and simulating far-field speech.
These techniques, however, did not improve the speech recognition performance, and we discarded them from our experimental setup to limit the amount of comparisons.

\subsection{Additional Generated Training Data}
\label{exp-setup-gos}
For Gronings, we investigate the effect of using additional generated training data obtained through self-training or via a TTS system.
This additional training data is generated on the basis of the remaining manually transcribed speech data we have available for Gronings.
Specifically, from this data we only use the audio recordings combined with the associated automatically generated transcriptions in the self-training procedure, while we only use the transcriptions of these recordings together with the associated synthetic speech generated using the TTS system during the synthetic speech procedure (explained below).
We did not use the speech data in combination with the associated manually generated transcriptions for training, since we are interested in the performance of the two aforementioned data augmentation techniques.
Note that for these experiments, we only use the smallest subset of manually transcribed speech training data (i.e.~24 minutes) to investigate the added benefit of generating a relatively large amount of additional fine-tuning data.

Inspired by \citet{iterative-pseudo}, we conduct three iterations of self-training to incrementally improve the quality of the generated transcriptions.
Specifically, we fine-tune an \texttt{XLS-R} teacher model on the 24-minute subset of Gronings as the first step. This model is then used to transcribe the remaining unlabeled portion of the original training data (i.e.~168 minutes). The combined data is then used to fine-tune a student model.
We use the new student model to transcribe another set of 168 minutes of unlabeled speech, and add this data to our training data, which now contains 24 minutes of original data and two times 168 minutes (i.e.~336 minutes) of data that was transcribed through self-training.
We then fine-tune another student model using the new training data (i.e.~24 + 336 minutes) and use it to transcribe an additional set of 336 minutes of unlabeled data to examine the effects of substantially increasing the training data.
Finally, we also add these data to our training data and fine-tune a final student model on the complete amount of training data (i.e.~24 + 336 + 336 minutes). Each of these student models is then evaluated on the test set.

\subsection{Synthetic speech}
\begin{figure*}[ht!]
    \centering
    \includegraphics[width=0.79\textwidth]{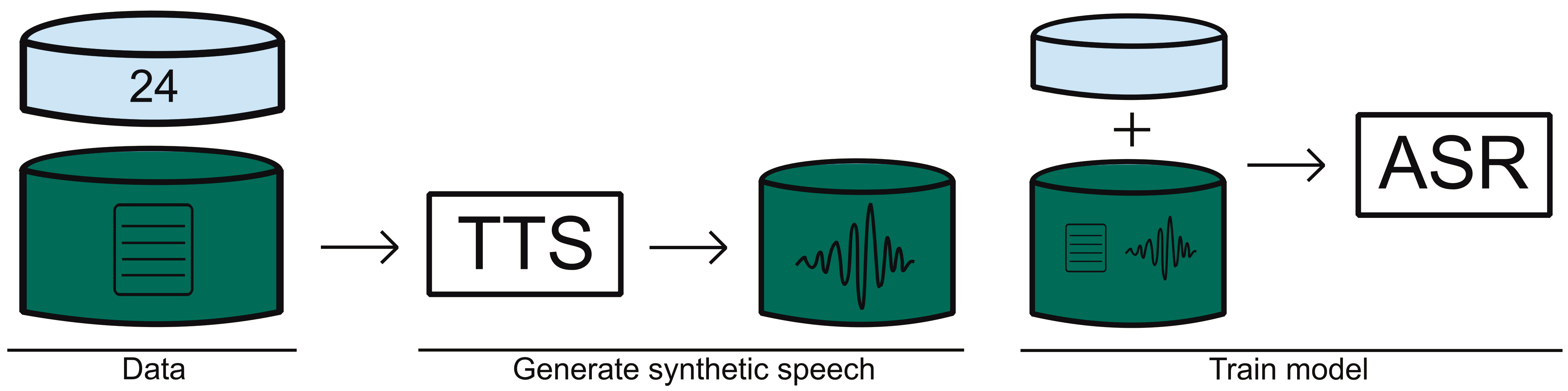}
    \caption{Visualization of the TTS-based approach, where synthetic speech is generated by an existing TTS model (trained on a separate two-hour single-speaker dataset), and new models are subsequently trained on both manually transcribed speech and synthetic speech.}
    \label{fig:method:tts}
\end{figure*}

In addition to transcribing unlabeled speech through self-training, we generate synthetic speech samples on the basis of the original transcriptions using an existing TTS system that was trained on about two hours of read speech from a single female speaker of the Hogelandsters variant of Gronings.
This system uses the FastSpeech 2 architecture \citep{fastspeech2}, and was previously developed for integration (pending) in  the online language documentation project on Gronings.\footnote{\url{https://woordwaark.nl}}
We use this existing TTS system to generate synthetic training data using the transcripts of the same sets of recordings that were used for the self-training experiments explained above.
To line up with the self-training models, we fine-tune three \texttt{XLS-R} models using different amounts of training data.
The first model is fine-tuned using the 24-minute  subset of manually transcribed speech supplemented with synthetic speech generated using the transcripts that correspond to the remaining 168 minutes of manually transcribed training data.
The second model is fine-tuned on the same subset augmented with the second set of 168 minutes of additional TTS-generated recordings (i.e.~based on the transcriptions of the second set of 168 minutes of training data also used in the self-training experiment described above).
We then augment the training data once more by adding synthetic speech samples using the transcripts from the final set of additional training data (i.e.~336 minutes), and fine-tune the \texttt{XLS-R} model on the complete amount of training data.
This approach is visualized in Figure~\ref{fig:method:tts}.

\section{Results}
We show the word error rates (WERs) for Gronings, West-Frisian, Besemah, and Nasal in Figure~\ref{fig:results:wers}.
The WERs for the development set are presented in Appendix~\ref{sec:appendix}.
For each of the languages, we observe a clear performance increase (i.e.~lower WERs) when the amount of manually transcribed training data becomes larger.
The WERs decrease between 30.1\% and 53.3\% when we use the complete set of training data (i.e.~192 minutes of manually transcribed speech data) instead of the 24-minute subset.
Importantly, Figure~\ref{fig:results:wers} also shows that self-training is beneficial for each of the languages. Student models improve over their teacher models in almost all cases. The improvement is particularly strong when the teacher model was based on a very small amount of data (i.e.~24 minutes) and ranges between 6.3\% and 13.9\%. 

\begin{figure*}[ht!]
     \centering
      \begin{subfigure}[b]{0.49\textwidth}
         \centering
         \includegraphics[width=\textwidth]{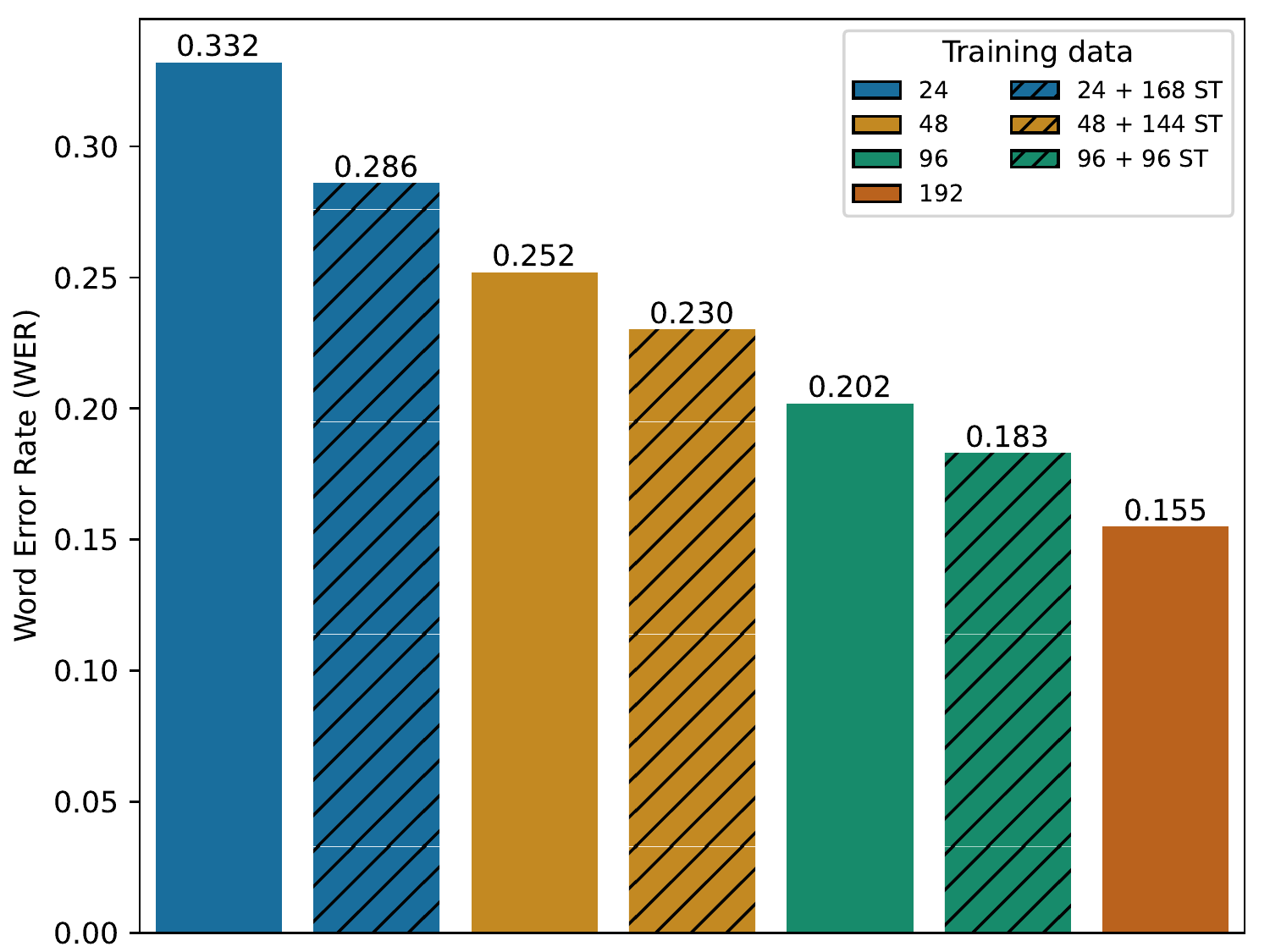}
         \caption{Results for the Gronings test set.}
         \label{fig:results:gos}
     \end{subfigure}
     \hfill
     \begin{subfigure}[b]{0.49\textwidth}
         \centering
         \includegraphics[width=\textwidth]{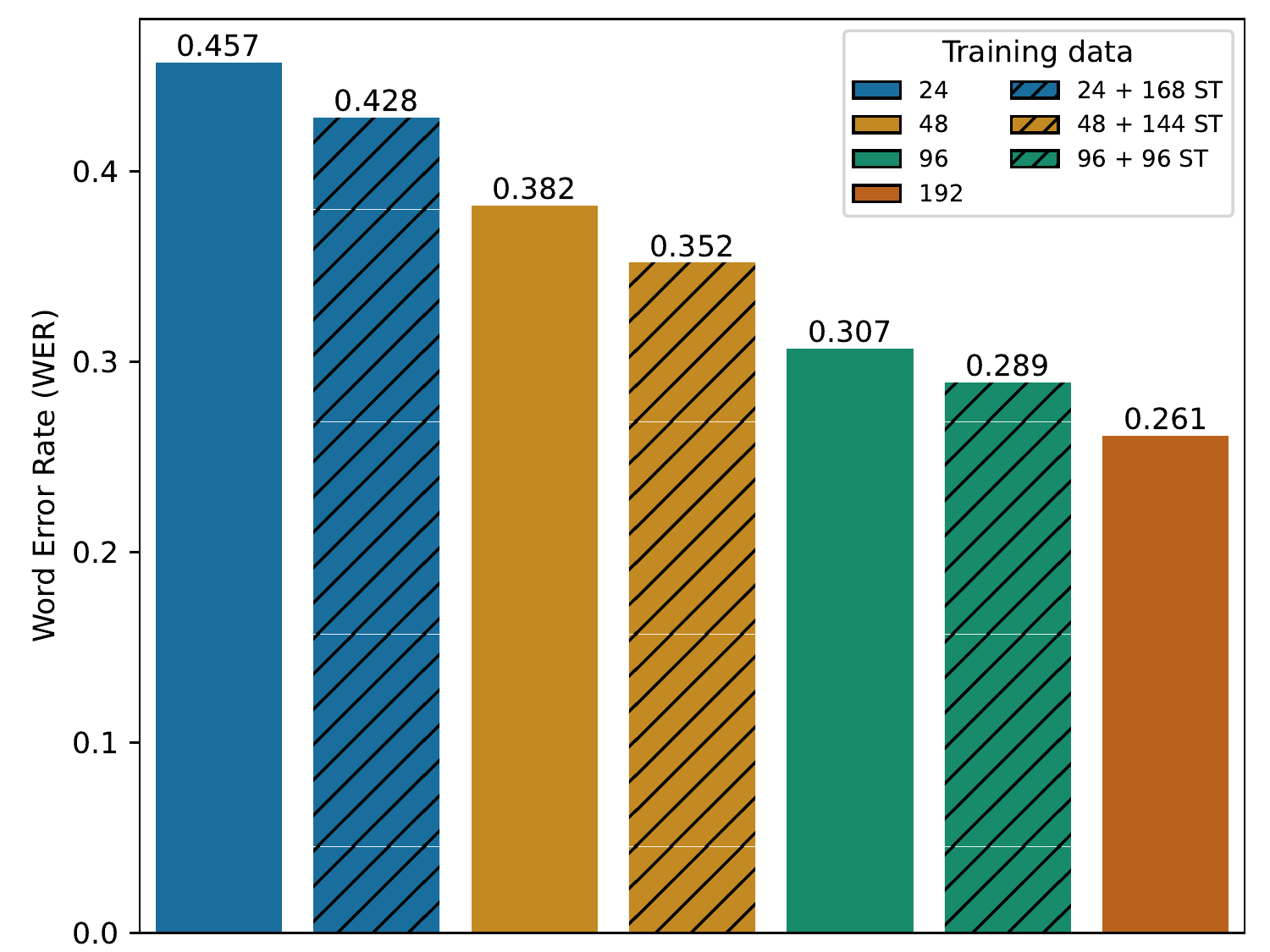}
         \caption{Results for the West-Frisian test set.}
         \label{fig:results:fry}
     \end{subfigure}
     \hfill
      \begin{subfigure}[b]{0.49\textwidth}
         \centering
         \includegraphics[width=\textwidth]{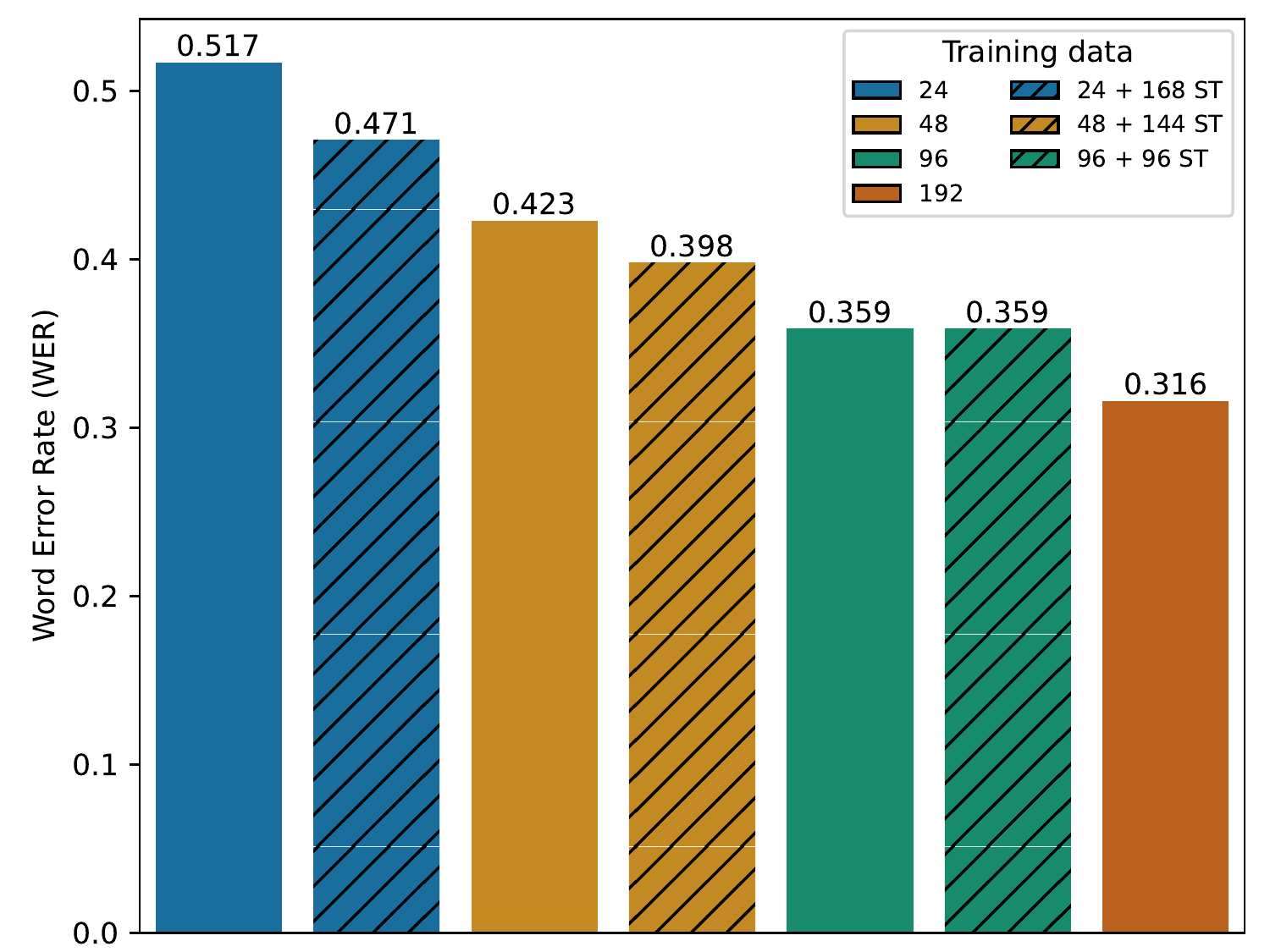}
         \caption{Results for the Besemah test set.}
         \label{fig:results:pse}
     \end{subfigure}
     \hfill
     \begin{subfigure}[b]{0.49\textwidth}
         \centering
         \includegraphics[width=\textwidth]{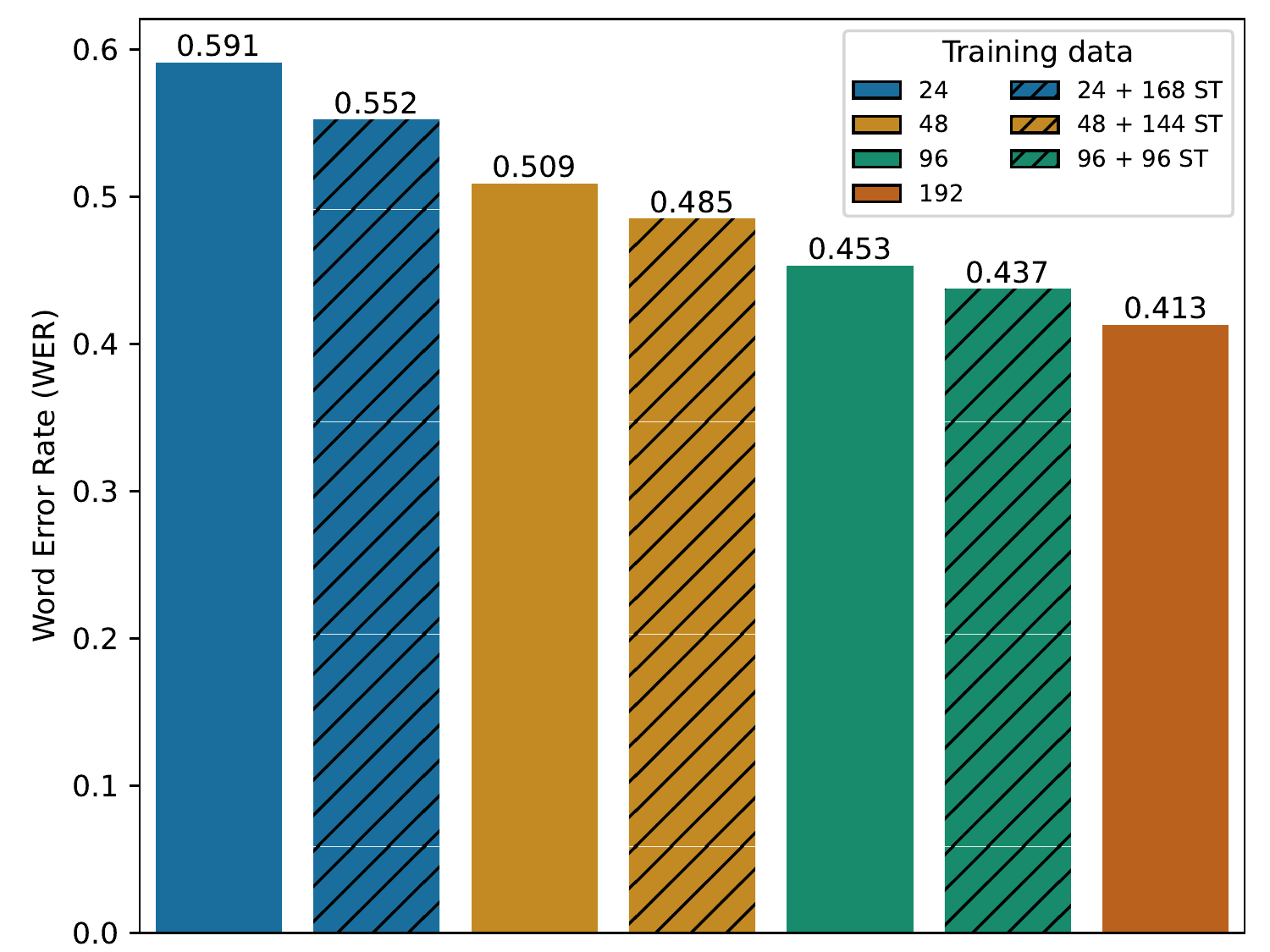}
         \caption{Results for the Nasal test set.}
         \label{fig:results:nsy}
     \end{subfigure}
     \caption{WERs for the test sets of Gronings, West-Frisian, Besemah, and Nasal using varying amounts of training data. Hatched bars indicate when additional training data generated by self-training (ST) was used.}
     \label{fig:results:wers}
\end{figure*}

\subsection{Further Pre-Training}
\begin{figure*}[ht!]
    \centering
    \includegraphics[width=0.8\textwidth]{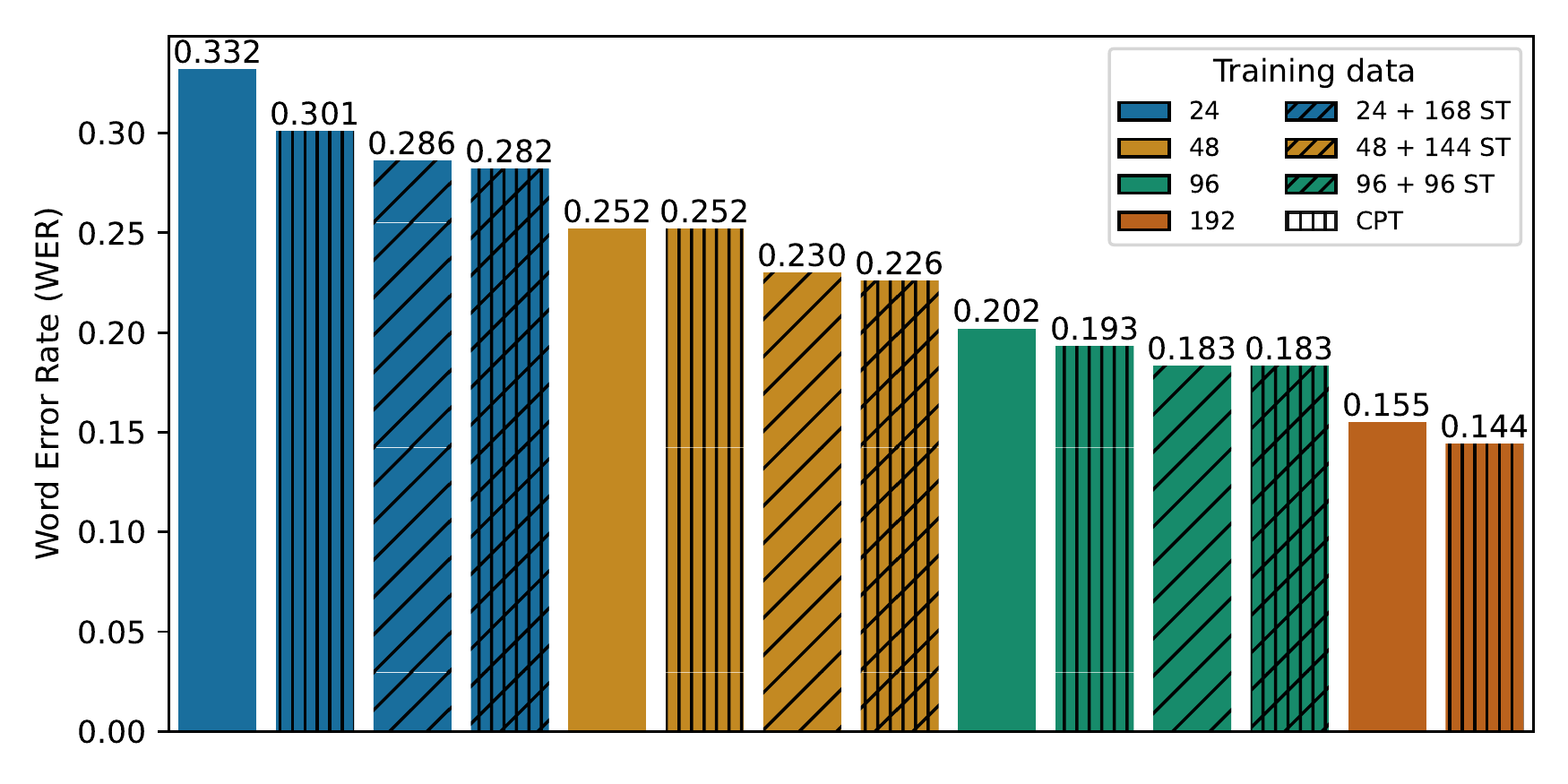}
    \caption{WERs for the test set of Gronings using an \texttt{XLS-R} model that was further pre-trained on Gronings (CPT: bars with vertical lines). Hatched bars indicate when additional training data generated by self-training (ST) was used. For comparison, the results using the model without further pre-training are shown as well (bars without vertical lines).}
    \label{fig:results:gos-cpt}
\end{figure*}

In Figure~\ref{fig:results:gos-cpt}, we show the fine-tuning results for varying amounts of training data (similar to those shown in Figure~\ref{fig:results:wers}) based on an \texttt{XLS-R} model that was further pre-trained on Gronings.
For comparison, this figure also shows the performance of the original fine-tuned models for Gronings. 
Pre-training generally results in a small increase in performance (up to a 9.3\% improvement) when only manually transcribed speech data was used to fine-tune the model.
Additionally, when a model was fine-tuned on data obtained using self-training, the performance gains were minimal (up to 1.7\% improvement).

\subsection{Additional Generated Training Data}
The effect of using additional augmented training data on ASR model performance is visualized in Figure~\ref{fig:results:gos-extra}.
To better evaluate these results, we also added the self-training results shown in Figure~\ref{fig:results:gos} to this figure.
Our results for self-training show that increasing the amount of automatically generated fine-tuning data is beneficial, albeit to a lesser extent than the benefit of using the first set of 168 minutes of speech with automatically generated transcriptions.
Nevertheless, the performance of the model fine-tuned using 24 minutes of manually transcribed speech data plus 672 minutes of speech data with automatically generated transcriptions yields a relative WER reduction of 20.5\% compared to the corresponding teacher model.
Consequently, its performance is close to the performance of the model fine-tuned on 48 minutes of manually transcribed speech data.

Figure~\ref{fig:results:gos-extra} also shows that an even greater performance gain, namely a WER reduction of 38.6\% relative to the model trained using 24 minutes of manually transcribed speech, can be achieved when using an existing TTS system to generate additional training data.\footnote{\citet{pivot-lang-2022} show that synthetic speech from a high-resource language TTS system may be used to generate additional training data for a low-resource language. We experimented with an existing Dutch TTS system to generate synthetic speech for Gronings, but this did not lead to improvements in performance.}
There is no clear benefit, however, of generating successively larger sets of synthetic speech.
Nevertheless, the performance of the model fine-tuned using 24 minutes of manually transcribed speech data plus 168 minutes of synthetic speech data generated using the TTS systems is almost identical to the performance of a model fine-tuned using 96 minutes of manually transcribed speech data. 

\subsection{Out-of-domain results}
\begin{figure*}[ht!]
     \centering
      \begin{subfigure}[b]{0.49\textwidth}
         \centering
         \includegraphics[width=\textwidth]{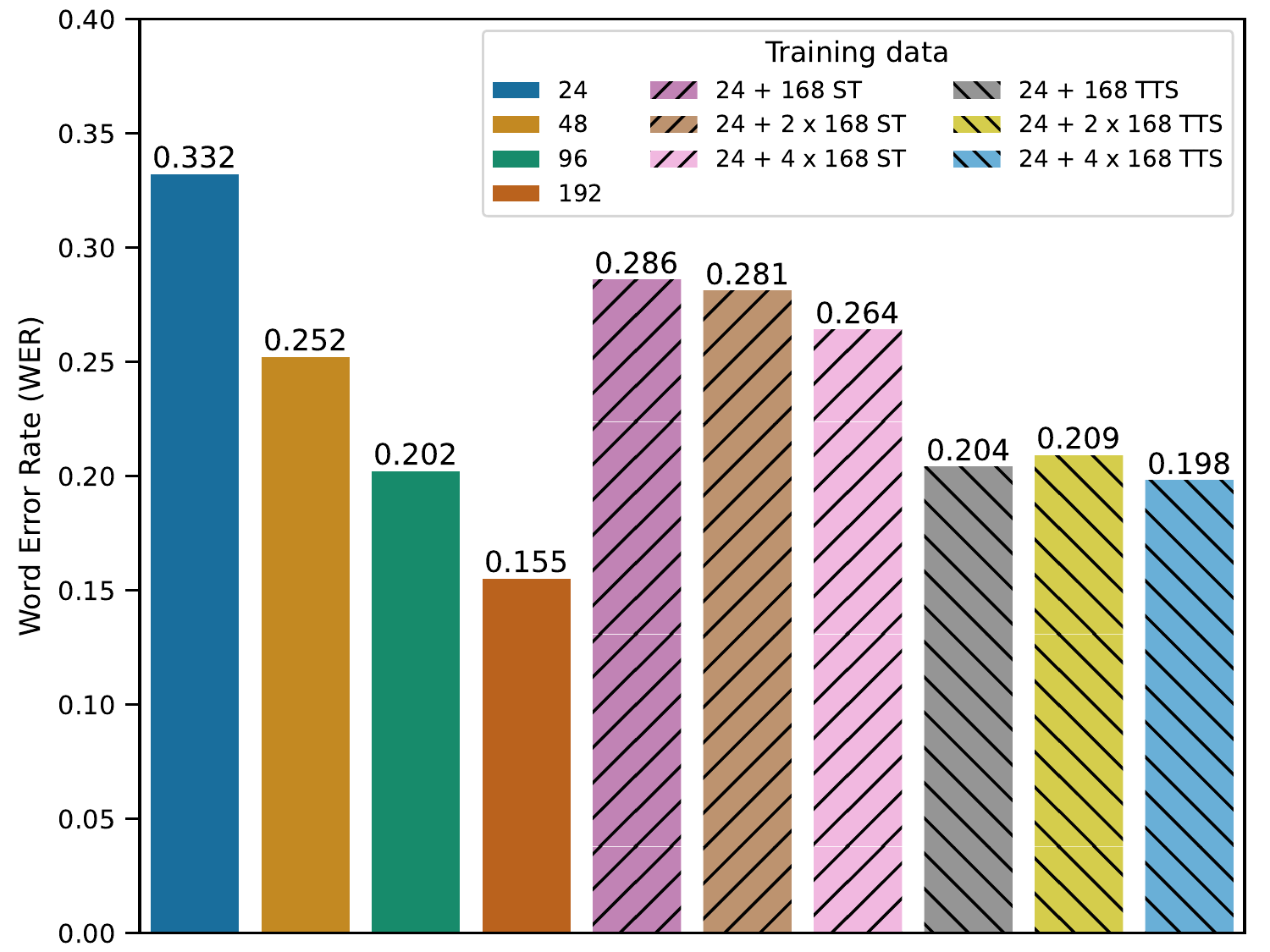}
         \caption{Results for the regular Gronings test set.}
         \label{fig:results:gos-extra}
     \end{subfigure}
     \hfill
     \begin{subfigure}[b]{0.49\textwidth}
         \centering
         \includegraphics[width=\textwidth]{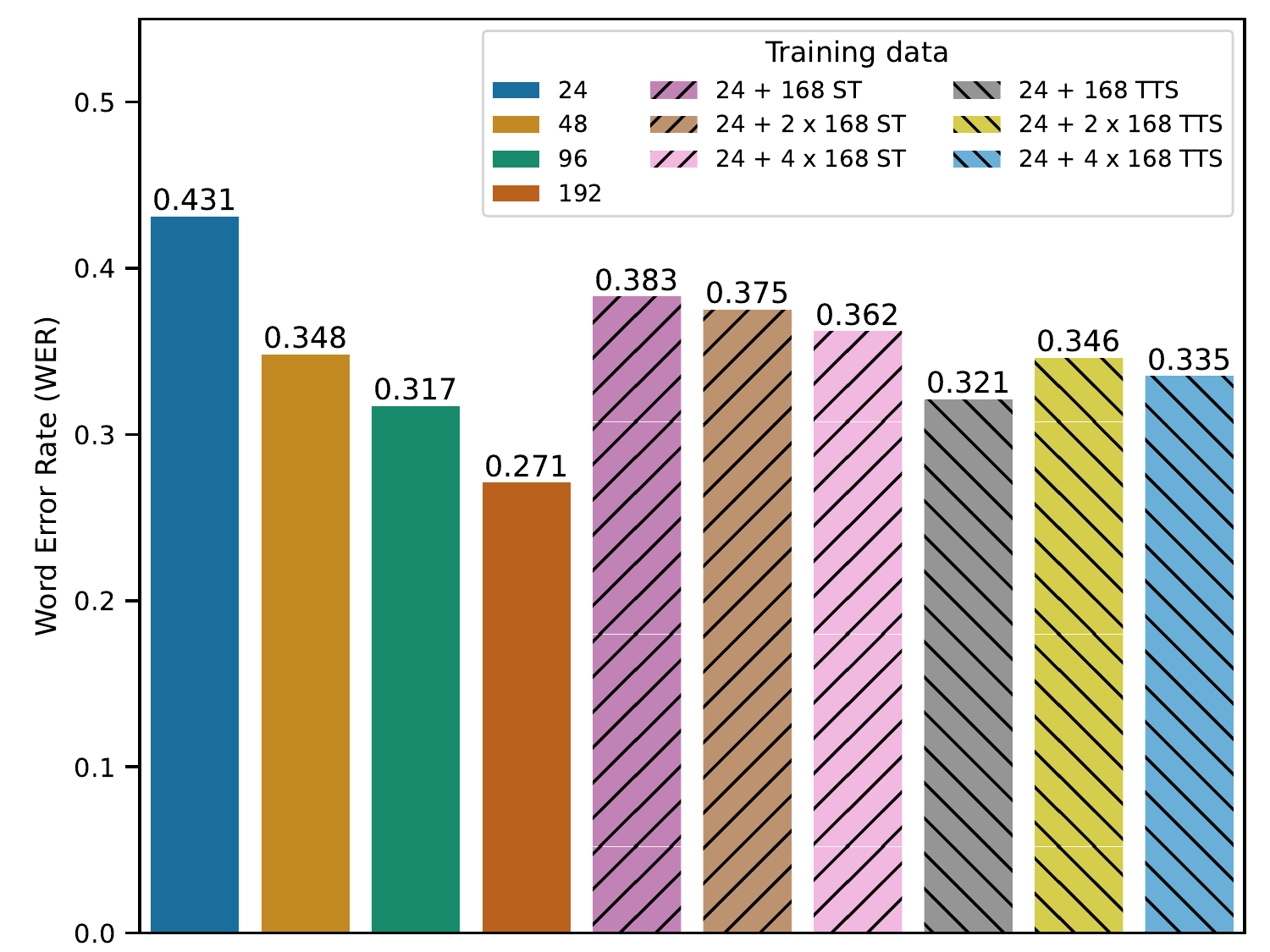}
         \caption{Results for the out-of-domain Gronings test set.}
         \label{fig:results:gos-extra-ood}
     \end{subfigure}
     \caption{WERs for the regular test set and out-of-domain test set of Gronings when additional training data generated by self-training (ST) or a text-to-speech system (TTS) was used.}
     \label{fig:results:wers-extra-ood}
\end{figure*}

The results presented in Figure~\ref{fig:results:gos-extra} might overestimate the model performance, as the speaker whose data was used for training the available TTS system was also included in the Gronings test set.
We therefore also report the fine-tuned model performance on an out-of-domain test set, which does not include any of the speakers that are included in the training data.
The results are shown in Figure~\ref{fig:results:gos-extra-ood}.
While the performance on the out-of-domain data is clearly worse compared to the original test set, the pattern of the results for the self-training approach remains similar (with a relative WER improvement of up to 16.0\%).
Furthermore, the benefit of augmenting the training data using a TTS system is still present, but it is less pronounced than before (with a WER improvement of up to 25.5\%).
Nevertheless, both data augmentation techniques still offer a substantial improvement in WER when the availability of manually transcribed training data is limited.

\section{Discussion and Conclusion}
We investigated whether data augmentation techniques are beneficial to improve the performance of ASR systems for four typologically different languages with a limited amount of real-world training data available.
We evaluated the performance of \texttt{XLS-R} models fine-tuned using varying amounts of training data, showing that the model performance generally improves (i.e.~resulting in lower WERs) when (more, in the case of self training) augmented training data is used.
The greatest performance gains across the four languages were observed when the amount of manually transcribed data used for fine-tuning was increased.
Nevertheless, we also observed substantial increases in model performance by augmenting very limited amounts of training data through self-training.
For Gronings, we found that fine-tuning a model on additional data obtained through iterative self-training performed almost as well as a model fine-tuned on double the amount of manually transcribed speech data.
Importantly, self-training only requires collecting additional unlabeled speech data, which is typically much easier to obtain than transcribed speech, making it a valuable approach for low-resource languages.

Moreover, using an existing TTS system for generating additional synthetic training data was likewise shown to be beneficial.
We observed that the benefit of augmenting the training data via the TTS system yielded larger performance gains (even on par with a model fine-tuned on four times the minimum amount of manually transcribed speech data we considered) than using the iterative self-training procedure.
However, in contrast to self-training, no beneficial effect was present when increasing the amount of generated data.
This pattern held true irrespective of using the general test set for evaluation or an out-of-domain test set instead.
While not many minority languages have a suitable TTS system available, generating speech data using such a system is very easy as it only requires written text.
Of course, our results also show that when the material is available to train a TTS system (i.e.~using audio recordings and associated transcriptions) it is likely better to use these resources directly for training the ASR system.  

While we showed the benefit of iterative self-training when a very small amount of training data is available, the benefit of supplying more and more self-trained training data was diminishing. 
Our result extends the findings for English by \citet{iterative-pseudo} to a new set of languages variants.
It is possible that the transcriptions generated by a specific teacher model in the self-training approach contain useful information, but that this is negated to a large extent by the generated errors of the model.
As teacher models fine-tuned on larger amounts of manually transcribed training data are expected to yield higher quality transcriptions (as shown in e.g., \citealt{san-etal-2022-automated}), the effect of generating more data might be more beneficial in these cases.
However, this should be investigated in future work.

When using the TTS system for augmenting our training data, we did not see a benefit of increasing the amount of generated synthetic speech.
As the additional training data represents data from a single speaker (as the TTS system was trained on the basis of data from a single speaker), the model might have been been overfitting to that specific speaker.
Future work, therefore, needs to investigate alternatives (or additions) to using a TTS system for generating additional training data.
For example, by investigating whether model performance can be improved using speaker adaptation methods or cross-lingual voice conversion (e.g., \citealt{rossenbach:20, baas22_interspeech}).

We found only minor performance gains when we fine-tuned the \texttt{XLS-R} model that was further pre-trained on Gronings (using all training and development data).
Specifically, self-training appeared to have greater performance gains than continuing pre-training (CPT), and combining CPT and self-training only marginally improved results.
Given the large computational cost of CPT as opposed to the two data augmentation methods, it is clear that CPT is not cost-effective.
It may be that CPT only yields appreciable performance gains once a sufficient amount of unlabeled audio can be obtained \citep[e.g.~200 hours of Ainu:][]{ainu2023}.
However, obtaining such a large amount of data for minority languages or language variants such as Gronings, Besemah, and Nasal is unlikely.
It is therefore important to further investigate how a limited amount of target language data can be used effectively for self-supervised pre-training.
For example, \citet{cpt-hr-2022} reported that using an additional 70-hour out-of-domain corpus alongside a 12-hour target corpus was crucial in improving performance.
Given that similar language regularization approaches have been effective for neural machine translation \citep[e.g.][]{neubig-hu-2018-rapid}, it may be possible that this strategy could also be beneficial for further pre-training in speech (e.g., using a 70-hour Indonesian speech corpus alongside the target four hour Besemah corpus).

In conclusion, our results show that data-augmentation techniques may serve as a cost-effective way to improve ASR performance for low-resource languages and variants.
While the performance of the four systems is not comparable to systems developed for high-resource languages, these systems may serve as a starting point for these language varieties.
We hope our experiments help further more inclusive speech technology for low-resource languages.

\section*{Limitations}
While we show a clear benefit of data augmentation when the amount of available training data is limited, the performance gain seems to be lower when a larger quantity of manually transcribed speech data is available.
Whether data augmentation is always beneficial is an open question. 

We did not measure the effect of sociolinguistic variables on the performance of the models.
A risk might be that especially for the models for Gronings, which were developed on the basis of speech data from only a few speakers, results might be negatively affected by differences in language background (such as speaking a different variety of Gronings, or being from a different social group).
We likewise did not measure the effect of non-linguistic variation (e.g., use of different microphones) on the performance of the models.
While \citet{BARTELDS2022101137} showed that wav2vec 2.0 representations are relatively unaffected by non-linguistic variation, we aim to further explore this in future work. 

Finally, we evaluated the effect of training data size and data augmentation on four different minority languages or language variants, each using a single test set.
Of course, using a different test set might have affected the results.
However, given that the pattern of results was similar across a range of language varieties we do not expect this difference to be large.

\section*{Ethics Statement}
Our paper evaluated various methods that could make developing automatic speech recognition systems more viable for languages where paired audio and transcriptions are difficult to obtain. 
In our experiments, we only used already publicly available data (West-Frisian) or data for which we have obtained informed consent for public release from the data custodians (Gronings, Besemah, Nasal). To make our findings as relevant as possible for other language projects, we minimized the amount of computing time used.

\section*{Acknowledgements}
The authors thank the Center for Information Technology of the University of Groningen for their support and for providing early access to the Habrok high performance computing cluster.
We also thank the community members of the four languages, and the three anonymous reviewers for their insightful feedback.

\bibliography{anthology,custom}
\bibliographystyle{acl_natbib}

\appendix
\onecolumn

\section{Results on Development Data}
\label{sec:appendix}

\begin{figure*}[ht!]
     \centering
      \begin{subfigure}[b]{0.49\textwidth}
         \centering
         \includegraphics[width=\textwidth]{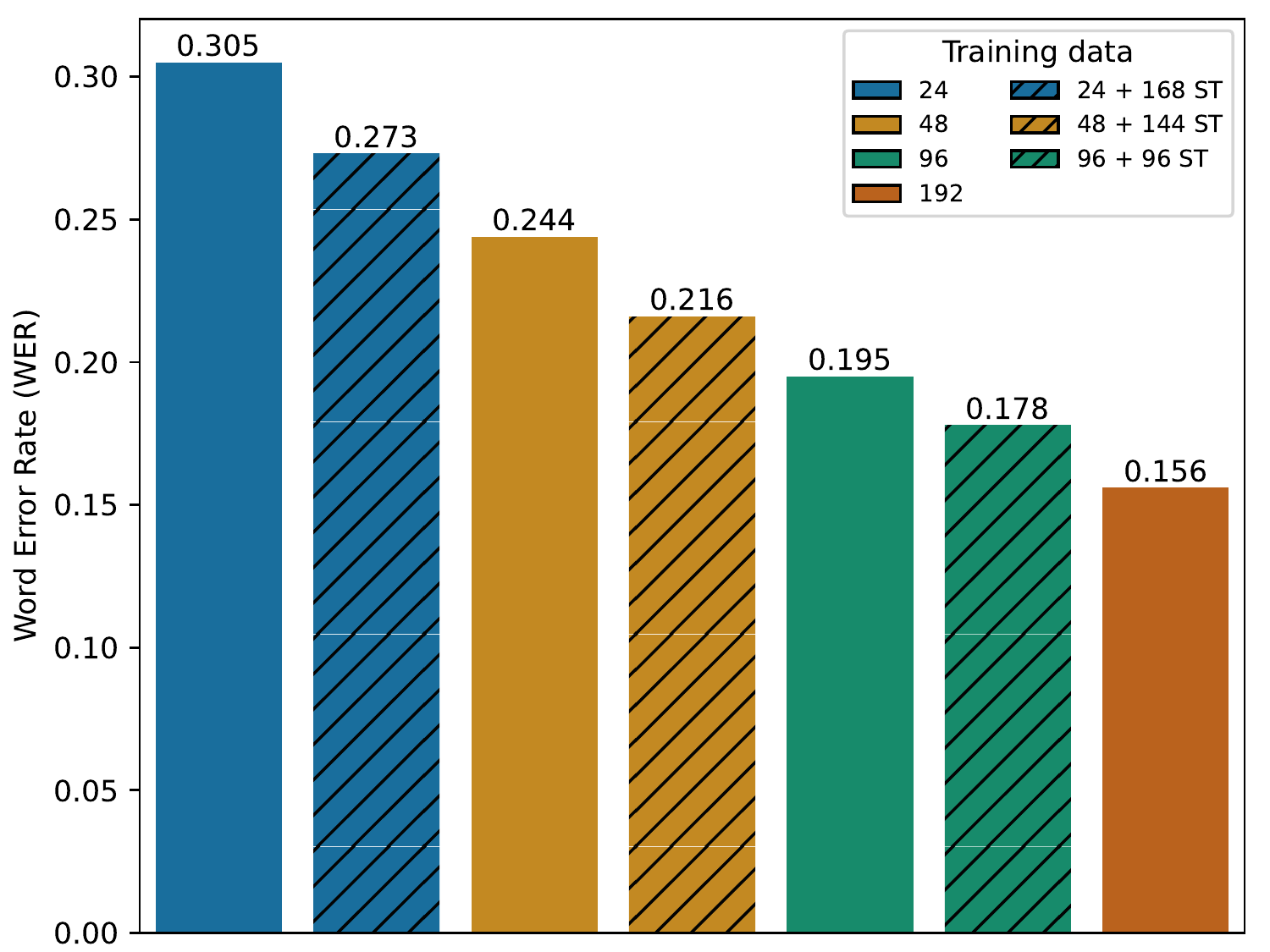}
         \caption{Results for the Gronings development set.}
         \label{fig:results:gos-dev}
     \end{subfigure}
     \hfill
     \begin{subfigure}[b]{0.49\textwidth}
         \centering
         \includegraphics[width=\textwidth]{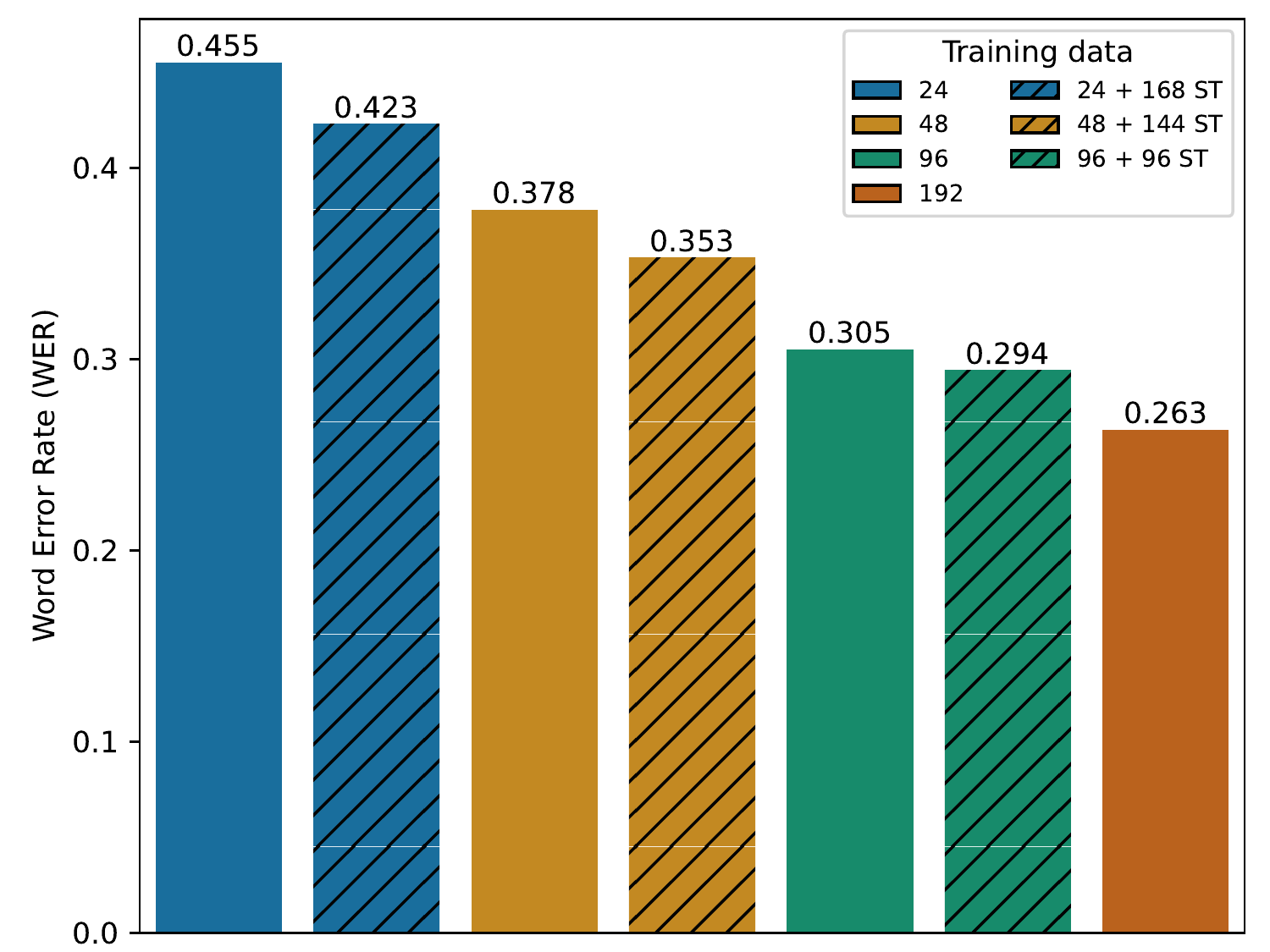}
         \caption{Results for the West-Frisian development set.}
         \label{fig:results:fry-dev}
     \end{subfigure}
     \hfill
      \begin{subfigure}[b]{0.49\textwidth}
         \centering
         \includegraphics[width=\textwidth]{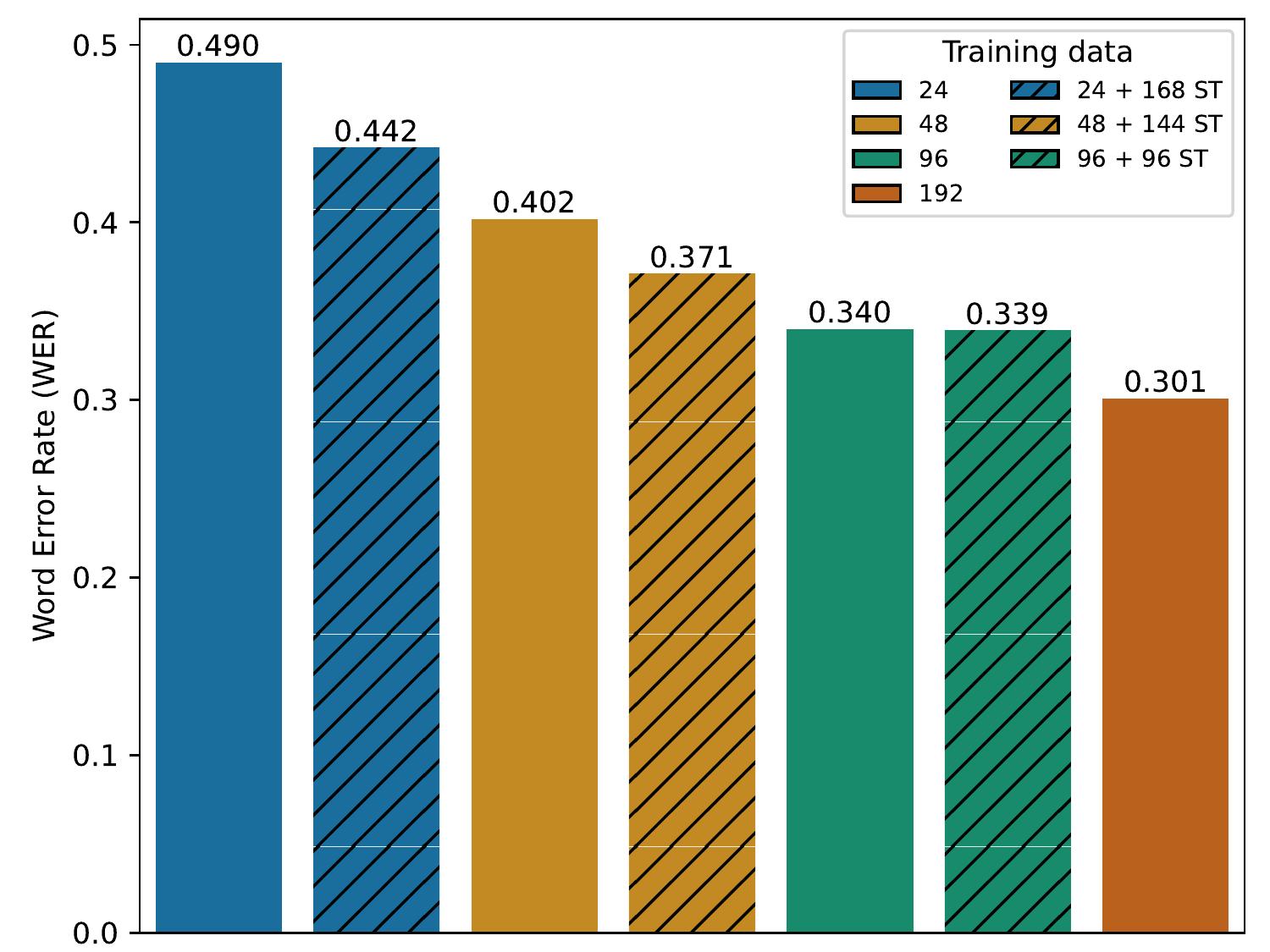}
         \caption{Results for the Besemah development set.}
         \label{fig:results:pse-dev}
     \end{subfigure}
     \hfill
     \begin{subfigure}[b]{0.49\textwidth}
         \centering
         \includegraphics[width=\textwidth]{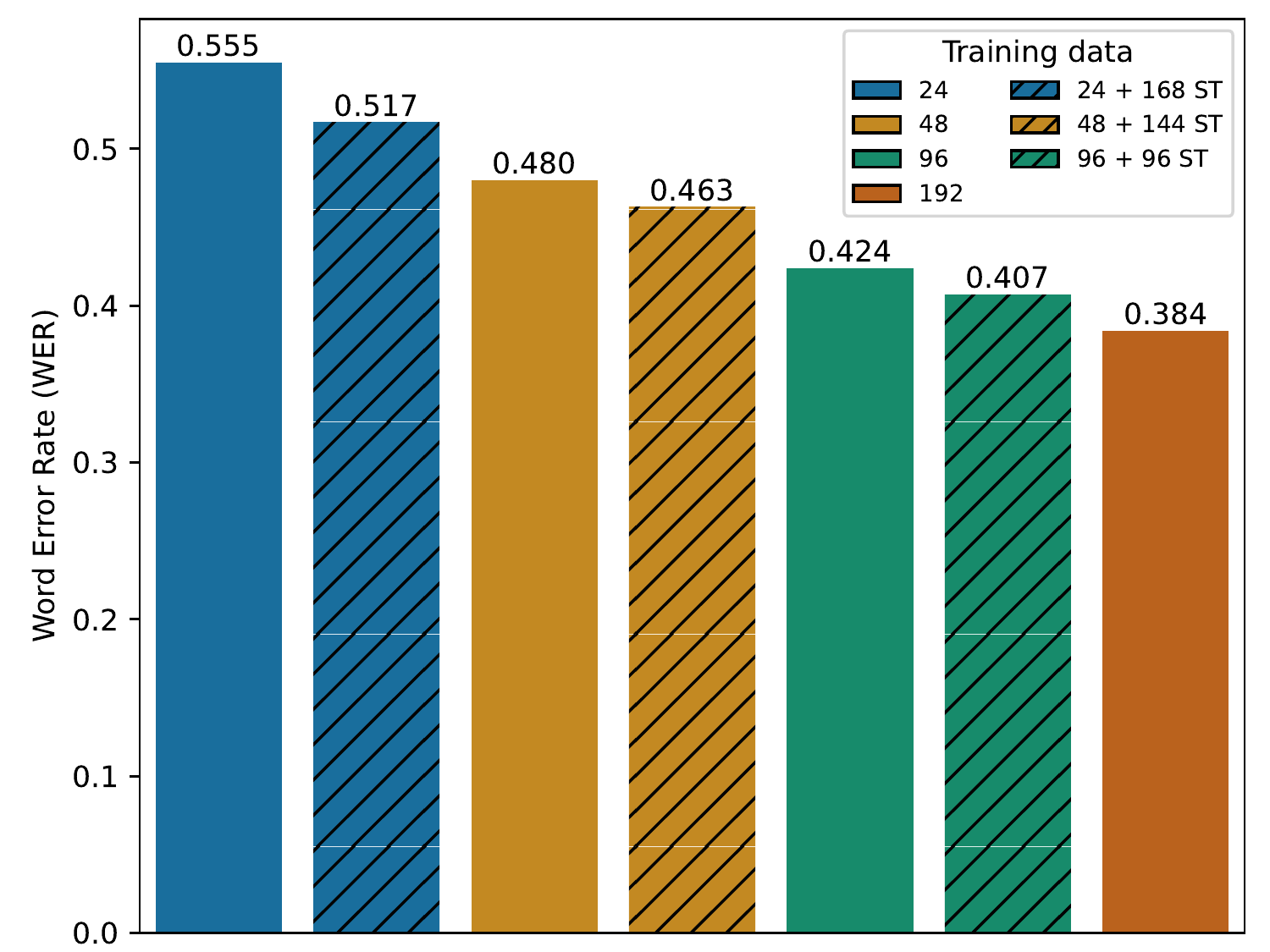}
         \caption{Results for the Nasal development set.}
         \label{fig:results:nsy-dev}
     \end{subfigure}
     \caption{WERs for the development sets of Gronings, West-Frisian, Besemah, and Nasal using varying amounts of training data. Hatched bars indicate when additional training data generated by self-training (ST) was used.}
     \label{fig:results:wers-dev}
\end{figure*}

\begin{figure*}[ht!]
    \centering
    \includegraphics[width=0.89\textwidth]{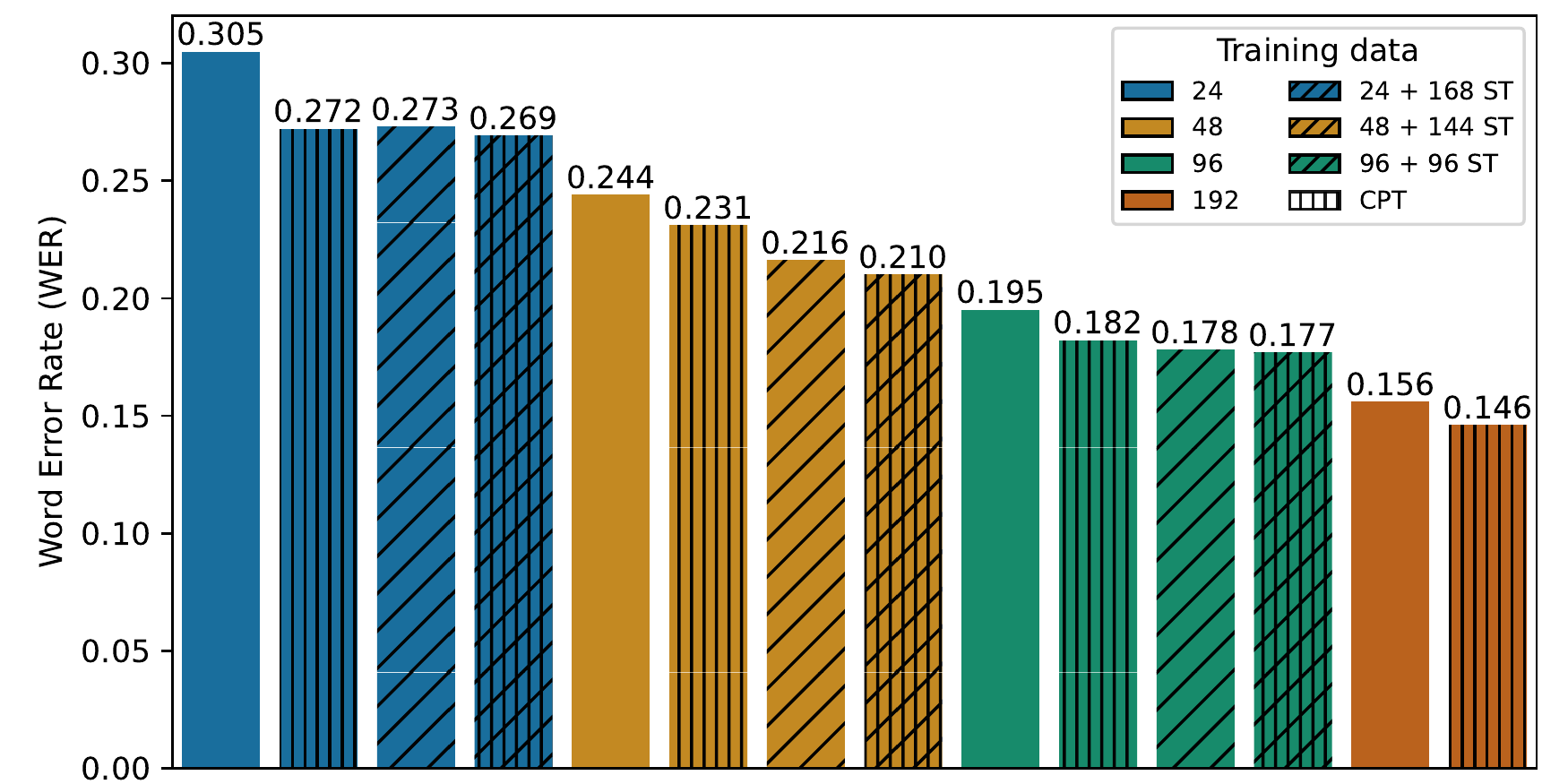}
    \caption{WERs for the development set of Gronings using an \texttt{XLS-R} model that was further pre-trained on Gronings (CPT: bars with vertical lines). Hatched bars indicate when additional training data generated by self-training (ST) was used. For comparison, the results using the model without further pre-training are shown as well (bars without vertical lines).}
    \label{fig:results:gos-cpt-dev}
\end{figure*}

\begin{figure*}[ht!]
    \centering
    \includegraphics[width=0.89\textwidth]{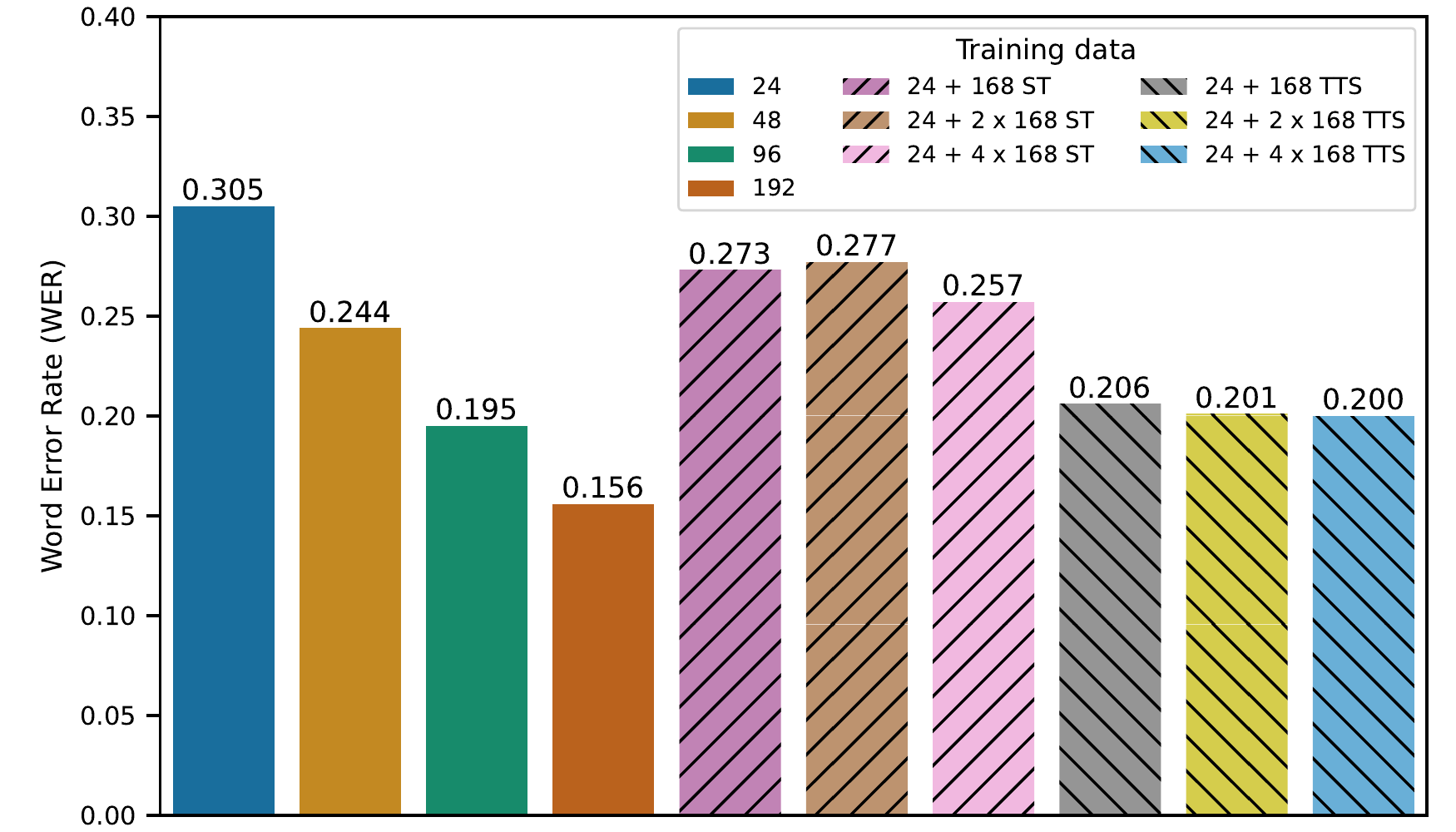}
    \caption{WERs for the development set when additional training data generated by self-training (ST) or a text-to-speech system (TTS) was used.}
    \label{fig:results:gos-extra-dev}
\end{figure*}

Figure~\ref{fig:results:wers-dev} shows the WERs for Gronings, West-Frisian, Besemah, and Nasal for the development set.
We show the fine-tuning results for varying amounts of training data using a model that was further pre-trained on Gronings in Figure~\ref{fig:results:gos-cpt-dev}.
Finally, the WERs in Figure~\ref{fig:results:gos-extra-dev} visualize the results for the development set of Gronings when additional training data generated by self-training (ST) or a text-to-speech system (TTS) was used.
Note that the pattern of these results is very similar to our findings for the test set.

\end{document}